\documentclass[letterpaper, 10 pt, conference]{ieeeconf}  %

\IEEEoverridecommandlockouts                              %

\usepackage{graphics} %
\usepackage{epsfig} %
\usepackage{amsmath} %
\usepackage{amssymb}  %
  
\usepackage{amssymb}
\usepackage{amsmath}
\usepackage{commath}
\usepackage{mathtools}

\usepackage{hyperref}
\usepackage{import}
\usepackage{paralist}
\usepackage{gensymb}
\usepackage{dsfont}
\usepackage{siunitx}
\usepackage[bottom]{footmisc}
\usepackage{framed}
\usepackage{balance} 
\usepackage{caption}
\usepackage[skins]{tcolorbox}  
\usepackage{subcaption}
\DeclareMathOperator*{\argmin}{arg\,min}
\hypersetup{
    colorlinks=true,
    linkcolor=black,
    citecolor=black,
    filecolor=black,
    urlcolor=black,
}
\DeclareCaptionFont{mysize}{\fontsize{8}{10}\selectfont}
\makeatletter
\newcommand*{\rom}[1]{\expandafter\@slowromancap\romannumeral #1@}
\makeatother

\title{\LARGE \bf
 Online DCM Trajectory Generation for Push Recovery \\ of Torque-Controlled Humanoid Robots
}

    \author{Milad Shafiee$^{\dagger, 1 }$, and Giulio Romualdi$^{\dagger, 1, 2}$, Stefano Dafarra$^{1, 2}$,  \\ Francisco Javier Andrade Chavez$^{1}$ and Daniele Pucci$^{1}$ %
\thanks{$\dagger$ Both authors equally contributed to the paper.}%
\thanks{$^{1}$ Dynamic Interaction Control, Italian Institute of Technology,
Genoa, Italy, {\tt\small (e-mail: name.surname@iit.it)}}
\thanks{$^{2}$ DIBRIS, University of Genoa, Genoa, Italy}
}

\begin{document}

\maketitle
\thispagestyle{empty}
\pagestyle{empty}

\begin{abstract}

We present a computationally efficient method for online planning of bipedal walking trajectories with push recovery. 
In particular, the proposed methodology fits control architectures where the Divergent-Component-of-Motion (DCM) is planned beforehand, and adds a \emph{step adapter} to adjust the planned trajectories and achieve push recovery. Assuming that the robot is in a single support state, the \emph{step adapter} generates new positions and timings for the next step. The \emph{step adapter} is active in single support phases only, but the proposed torque-control architecture considers double support phases too. The key idea for the design of the \emph{step adapter} is to impose both initial and final DCM step values using an exponential interpolation of the time varying ZMP trajectory.
This allows us to cast the push recovery problem as
a Quadratic Programming (QP) one, and to solve it online with state-of-the-art optimisers. 
The  overall  approach  is  validated  with simulations  of  the  torque-controlled  33  kg  humanoid
robot iCub.
Results show that the proposed strategy prevents the humanoid robot from falling while walking at 0.28 m/s and  pushed with external forces up to 150 Newton for 0.05 seconds.

\end{abstract}

\section{Introduction}
Bipedal locomotion of humanoid robots remains an active research domain despite decades of studies in the subject. The complexity of the robot dynamics, the unpredictability of its surrounding environment, and the low efficiency of the robot actuation system are only a few issues that complexify the achievement of robust robot locomotion. In the large variety of robot controllers implemented for bipedal locomotion, the Divergent-Component-of-Motion (DCM) is an ubiquitous concept used for generating walking patterns. 
In case of unforeseen disturbances acting on the robot, it is  necessary to adjust the generated pre-planned patterns from preventing the robot to fall.  This paper contributes towards this direction by presenting algorithms and architectures for achieving bipedal humanoid robot with real-time push recovery features. 

\par

A popular approach to humanoid robot control during the DARPA Robotics Challenge was to define a hierarchical architecture consisting of several layers~\cite{feng2015optimization}. Each layer generates references for the layer below by processing inputs from the robot, the environment, and the outputs of the previous layer. From top to bottom, these layers are here called: \emph{trajectory optimization}, \emph{simplified model control}, and \emph{whole-body quadratic programming (QP) control}. %

The \emph{trajectory optimization} layer often generates  desired foothold locations by means of optimization techniques. To do so, both kinematic and dynamical robot models can be used~\cite{dai2014whole,herzog2015trajectory}. When solving the optimization problem associated with the \emph{trajectory optimization} layer, computational time may be a concern especially when the robot surrounding environment is not structured.
There are cases, however, where simplifying assumptions on the robot environment can be made, thus reducing the associated computational time. For instance, flat terrain allows one to view the robot as a simple unicycle \cite{PascalHandbook,flavigne2010reactive}, which enables quick solutions to the optimization problem for the walking pattern generation~\cite{8594277}.

\begin{figure}[t]
  \centering
      \begin{subfigure}[b]{0.32\columnwidth}
        \centering
        \includegraphics[width=\textwidth]{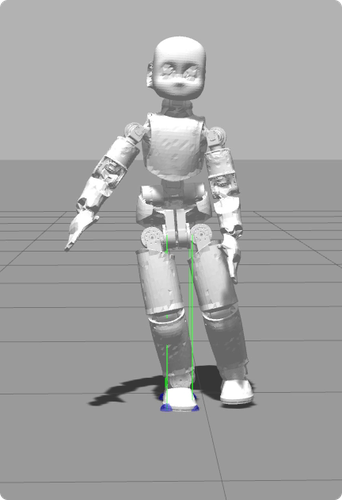}
    \end{subfigure}
    \begin{subfigure}[b]{0.32\columnwidth}
        \centering
        \includegraphics[width=\textwidth]{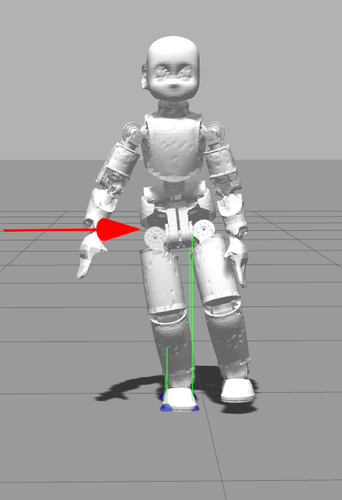}
    \end{subfigure}
    \begin{subfigure}[b]{0.32\columnwidth}
        \centering
        \includegraphics[width=\textwidth]{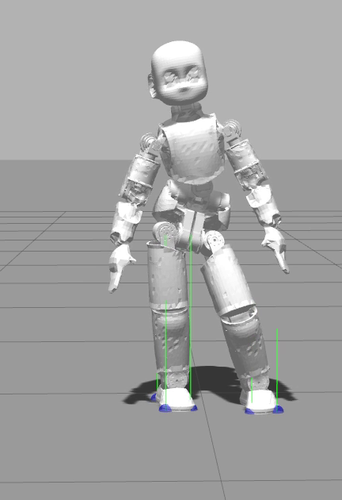}
    \end{subfigure}
  \caption{iCub reacts to the application of an external force (the red arrow). \label{fig:icub}}
\end{figure}

\par
The \emph{simplified model control} layer is in charge of finding feasible center-of-mass (CoM) trajectories and is often based on simplified  models, such as the Linear Inverted Pendulum Model (LIPM) \cite{Kajita2001} and the Capture Point (CP) \cite{Pratt2006,Hof2008}. These models have become very popular after the introduction of the Zero Moment Point (ZMP) as a stability criterion~\cite{Vukobratovic1969}. %
Another model that is often exploited in the \emph{simplified model control} layer is the Divergent Component of Motion (DCM) \cite{Englsberger2015}. 
The DCM  can be viewed as the extension of the capture point (CP) concept to the three dimensional case.

The \emph{whole-body QP control} layer generates robot's joint torques depending on the available control modes of the underlying robot. These outputs aim at stabilizing the references generated by the previous layers. 
It uses whole-body kinematic and/or dynamical models, and very often instantaneous optimization techniques, namely, no MPC methods are here employed. Furthermore, the associated optimisation problem is often framed as a hierarchical stack-of-tasks, with strict or weighted hierarchies \cite{Stephens2010,nava16}.
\par
\par

In recent years, several attempts have been made for considering robot state feedback for footprints planning. The new foothold locations can be optimized  assuming constant step timing \cite{feng2016robust,Shafiee-Ashtiani2017,joe2018balance} or adaptive step duration \cite{khadiv2016step,griffin2017walking}.
These methods have the drawback to consider a constant Zero Moment Point (ZMP) during the single support phase. Attempts at relaxing this assumption involve the control of the Centroidal Momentum Pivot (CMP) position \cite{jeong2019robust1}, but the pre-planned ZMP is still not used during push recovery. The possibility of having a non-constant pre-planned ZMP during single support is fundamental for achieving toe-off motion: it indeed contributes towards humanoid robots  with energy efficient 
and human-like  walking \cite{kuo2002energetics,ogura2006human,Englsberger2014}.

This paper extends and encompasses the control architecture \cite{8594277,romualdi2018benchmarking} by complementing it with the \emph{push recovery} feature. 
In particular, the proposed methodology fits control architectures where the DCM is planned beforehand, and adds a \emph{step adapter} to adjust the planned trajectories and achieve push recovery. Assuming the robot in a single support state, the \emph{step adapter} generates new positions and timings for the next step. The \emph{step adapter} is active in single support phases only, but the proposed torque-control architecture considers double support phases too. The key idea for the design of the \emph{step adapter} is to impose both initial and final DCM step values using an exponential interpolation of the time varying ZMP trajectory.
This allows us to cast the push recovery problem as
a QP one, and to solve it online with state-of-the-art optimisers. 
The approach  is  validated  with simulations  of  the  torque-controlled  $\SI{33}{\kilo \gram}$  humanoid
robot iCub.
Results show that the proposed strategy prevents the  robot from falling while walking at $\SI{0.28}{\meter \per \second}$ and  pushed with external forces up to 150 Newton for 0.05 seconds
applied at the robot pelvis.
\par
The paper is organized as follows. Sec.~\ref{sec:BACKGROUND} introduces notation, the humanoid robot model, and some simplified models used for locomotion.
Sec.~\ref{sec:ARCHITETCURE} describes each layer of the control
architecture, namely the trajectory optimization, the simplified model control and the whole-body QP torque control layer. Sec.~\ref{sec:stepAdaptation} details the the step timing and position adaptation algorithm while Sec.~\ref{sec:RESULTS} presents the simulation results  on the iCub humanoid robot \cite{Natale2017} in push recovery scenarios. Finally, Sec.~\ref{sec:CONCLUSION} concludes the paper.

\section{Background}
\label{sec:BACKGROUND}
\subsection{Notation}
\begin{itemize}
\item $I_n$ and $0_n$ denote the $n \times n$ identity and zero matrices;
\item $\mathcal{I}$ denotes an inertial frame;
\item $\prescript{\mathcal{A}}{}{p}_\mathcal{C}\in \mathbb{R}^3$ is the vector from  the origin of $\mathcal{A}$ to the origin of  $\mathcal{C}$ expressed with respect to the  $\mathcal{A}$ orientation;
\item given $\prescript{\mathcal{A}}{}{p}_\mathcal{C}$ and $\prescript{\mathcal{B}}{}{p}_\mathcal{C}$,  $\prescript{\mathcal{A}}{}{p}_\mathcal{C} = \prescript{\mathcal{A}}{}{R}_\mathcal{B} \prescript{\mathcal{B}}{}{p}_\mathcal{C} + \prescript{\mathcal{A}}{}{p}_\mathcal{B}= \prescript{\mathcal{A}}{}{H}_\mathcal{B} \prescript{\mathcal{B}}{}{p}_\mathcal{C}$, where $\prescript{\mathcal{A}}{}{H}_\mathcal{B}$ is the homogeneous transformations and $\prescript{\mathcal{A}}{}{R}_\mathcal{B} \in SO(3)$ is the rotation matrix; 
\item given $w \in \mathbb{R}^3$ the \emph{hat operator} is $\hat{.} : \mathbb{R}^n \to \mathfrak{so}(3)$, where $\mathfrak{so}(3)$ is the set of skew-symmetric matrices and $\hat{x}y = x \times y$. $\times$ is the cross product operator in $\mathbb{R}^3$, in this paper the hat operator is also indicated by $S(.)$;
\item given  $W\in \mathfrak{so}(3)$ the \emph{vee operator} is $.^\vee : \mathfrak{so}(3) \to \mathbb{R}^3$;
\item $\prescript{\mathcal{A}}{}{\omega}_\mathcal{B} \in \mathbb{R}^3$ denotes the angular velocity between the frame $\mathcal{B}$ and the frame $\mathcal{A}$ expressed in the frame $\mathcal{A}$;
\item the subscripts $\mathcal{T}$, $\mathcal{LF}$, $\mathcal{RF}$ and $\mathcal{C}$ indicates the frames attached to the torso, left foot, right foot and CoM;
\item For the sake of clarity, the prescript $\mathcal{I}$ will be omitted.
\end{itemize}

\subsection{Humanoid Robot Model}
A humanoid robot is modelled as a floating base multi-body system composed of $n+1$ links connected by $n$ joints with one degree of freedom. Since none of the robot frames has an a priori pose with respect to (w.r.t.) the inertial frame $\mathcal{I}$, the robot configuration is completely defined by considering both the joint positions $s$ and the homogeneous transformation from the inertial frame to the robot frame (i.e. called base frame $\mathcal{B}$). 
In details, the configuration of the robot can be uniquely determined by the a triplet $q = (\prescript{\mathcal{I}}{}{p}_\mathcal{B}, \prescript{\mathcal{I}}{}{R}_\mathcal{B}, s) \in  \mathbb{R}^3 \times SO(3) \times \mathbb{R}^n$. 
The velocity of the floating system is represented by the triplet $ \nu = (\prescript{\mathcal{I}}{}{v}_\mathcal{B}, \prescript{\mathcal{I}}{}{\omega}_\mathcal{B}, \dot{s})$. %
Given a frame attached to a link of the floating base system, its position and orientation w.r.t. the inertial frame is uniquely identified by an homogeneous transformation, $\prescript{\mathcal{I}}{}{H}_\mathcal{A} \in SE(3)$. 
\par 
Similarly, the frame velocity w.r.t. the inertial frame is uniquely identified by the vector $\prescript{}{}{\mathrm{v}}_\mathcal{A} = \begin{bmatrix} \prescript{}{}{v}_\mathcal{A} ^ \top &  \prescript{}{}{\omega}_\mathcal{A}^ \top \end{bmatrix}^ \top$. The function that maps $\nu$ to the twist $\prescript{}{}{\mathrm{v}}_\mathcal{A}$ is linear and its matrix representation is the well known Jacobian matrix $J_\mathcal{A}(q)$:
\begin{equation}
    \label{eq:frame_velocity}
    \prescript{}{}{\mathrm{v}}_\mathcal{A} = J_\mathcal{A}(q) \nu.
\end{equation}
For a floating based system, the Jacobian can be split into two submatrices, one multiplies the base velocity while the other the joint velocities.
From \eqref{eq:frame_velocity}, one has:
\begin{equation}
    \prescript{}{}{\dot{\mathrm{v}}}_\mathcal{A} = J_\mathcal{A}(q) \dot{\nu} + \dot{J}_\mathcal{A}(q, \nu) \nu.
\end{equation}
The dynamics of the floating base system can be described by the Euler-Poincar\'e equation \cite{Marsden2010}:
\begin{equation}
\label{eq:robot_dynamic}
    M(q) \dot{\nu} + h(q, \nu) = B \tau + \sum_{k = 1}^{n_c} J^\top_{\mathcal{C}_k}(q) f_k,
\end{equation}
where $M(q) \in \mathbb{R} ^{(n + 6) \times(n + 6)}$ represents the mass matrix, $h(q, \nu) \in \mathbb{R} ^{n + 6 }$ is the Coriolis, the centrifugal term and the gravitational term. On the right-hand side, $B$ is a selector matrix, $\tau \in \mathbb{R}^n$ is the vector containing the joint torques and $f_k \in \mathbb{R}^6$ is a vector containing the coordinates of the contact wrench.
$n_c$ indicates the number of contact wrenches. Henceforth, we assume that at least one of the links is in contact with the environment, i.e. $n_c \ge 1$.
\par
The centroidal momentum of the robot $h$ is the aggregate linear and angular momentum of each link of the robot referred to the robot center of mass (CoM). The centroidal momentum of the robot is related to the robot velocity $\nu$ through the centroidal momentum matrix $A_g(q)$ as \cite{orin2008centroidal}
\begin{equation}
\label{eq:centroidal_momentum_matrix}
    h = A_g(q) \nu.
\end{equation}
The centroidal momentum time derivative is given by:
\begin{equation}
\label{eq:centroidal_momentum_matrix_rate_of_change}
    \dot{h} = A_g(q) \dot{\nu} + \dot{A}_g(q, \nu) \nu,
\end{equation}
where $\dot{A}_g(q, \nu) \nu$ is called centroidal bias acceleration.
It is worth to recall that the rate of change of the centroidal momentum of the robots can be also expressed by using the external contact wrenches acting on the system as \cite{nava16}
\begin{equation}
\label{eq:centroidal_momentum_dynamics}
    \dot{h} = \sum_{k = 1}^{n_c}\begin{bmatrix}
      I_3 & 0_3 \\
      S(p_{\mathcal{C}_k} - p_{\mathcal{C}}) & I_3 
    \end{bmatrix} f_k + mg.
\end{equation}

\subsection{Simplified models}
\label{sec:simplified_models}
The motion of the robot can be \emph{viewed} considering  the linear momentum dynamics only
\begin{equation}
\label{eq:3d-lipm}
    \dot{h}_l = m \ddot{x} = \sum_{k = 1}^{n_c} f_k^l + m g, 
\end{equation}
where $x \in \mathbb{R}^3$ represents the position of the CoM and $f_k^l$ the linear component of the external force acting on the system and $\dot{h}_l$ the rate of change of the linear centroidal momentum. 
\par
Now, the DCM is given by  \cite{Englsberger2015}:
\begin{equation}
    \label{eq:dcm}
\xi = x +  b \dot{x},
\end{equation}
where $b$, in case of constant CoM height $z_0$, is the pendulum time constant, i.e. $b = \sqrt{z_0/g}$.
 Using \eqref{eq:3d-lipm} and \eqref{eq:dcm}, the DCM time derivative is given by:
\begin{equation}
  \label{eq:dcm_dynamics_no_vrp}
  \dot{\xi} = \frac{1}{b}(\xi - x) + \frac{b}{m} \sum_{k = 1}^{n_c} f_k^l + b g.
\end{equation}
By introducing the virtual repellent point (VRP) as 
\begin{equation}
    r ^{vrp} = x - \frac{b^2}{m} \left(\sum_{k = 1}^{n_c} f_k^l  + m g  \right) = x - \frac{\dot{h}_l b^2}{m},
\end{equation}
Eq.~\eqref{eq:dcm_dynamics_no_vrp} can be simplified as:
\begin{equation}
  \label{eq:dcm_dynamics}
  \dot{\xi} = \frac{1}{b}(\xi - r ^{vrp}).
\end{equation}
\begin{figure*}[!t]
  \centering
\includegraphics[scale=1]{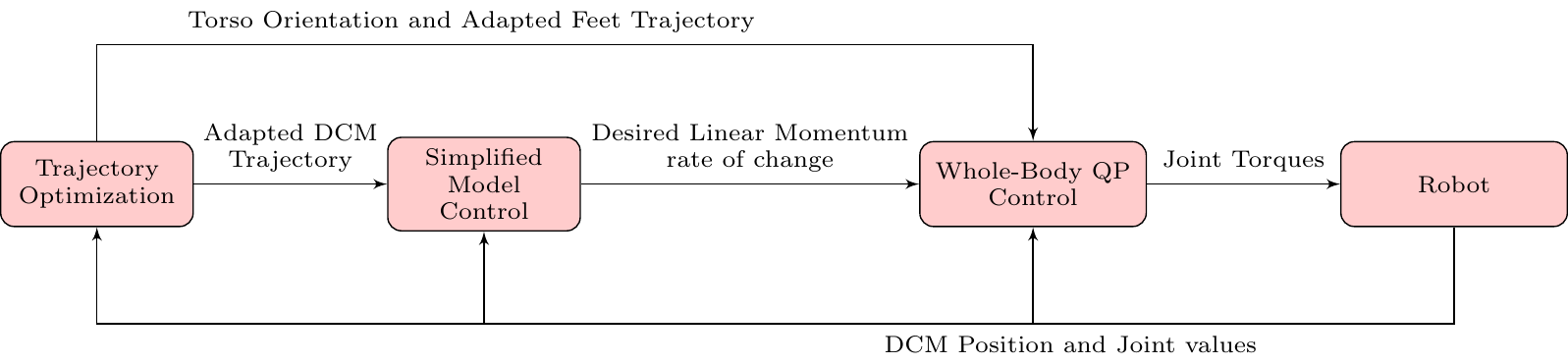}
  \caption{The control architecture is composed of three layers: the \emph{trajectory optimization}, the \emph{simplified model control}, and the \emph{whole-body QP control}.
  \label{fig:controller_architecture}}
\end{figure*}

\section{Torque-control DCM based architecture}
\label{sec:ARCHITETCURE}

This section summarizes a torque-control DCM based architecture for humanoid robots walking without the \emph{push recovery} feature -- see Fig.~\ref{fig:controller_architecture}. The \emph{push recovery} feature is added using the algorithm presented in Sec. \ref{sec:stepAdaptation}. %

The control architecture is composed of three main layers.
The first is the \emph{trajectory optimization}, whose main purpose is to generate desired footstep positions and orientation and also the desired DCM trajectory using the robot state. The second layer employs \emph{simplified robot models} to track the desired DCM trajectory. The third  layer is given by the \emph{whole-body QP} block. It has the purpose of ensuring the tracking of the desired feet positions and orientations and also the CoM acceleration by generating  joint torques. 

\subsection{Trajectory Optimization layer} \label{sec:trajectoryOptimization}
The objective of the trajectory optimization layer is to evaluate the desired feet and DCM trajectories. In this work, we divide the trajectory optimization problem into two sub-tasks. The former aims to generate the nominal footprints and DCM trajectory without taking into account the current status of the robot. Then the step adjuster module adjusts the aforementioned quantities by considering the error between the nominal and current DCM of the robot.
\par
In the trajectory optimization layer, we assume that the humanoid robot walks with a constant height between the CoM and the stance foot. This assumption allows us to simplify the DCM planning problem by considering only the projection of the DCM on the walking surface, while the height of the DCM remains constant and equal to the CoM height. Furthermore, by assuming a constant centroidal angular momentum, the projection of the VRP on the walking surface coincides with the ZMP \cite{Englsberger2014}. As a consequence, in this section we consider only the DCM planar dynamics.

Before generating the desired nominal trajectories, the nominal footstep positions have to be planned. For this, the humanoid robot is approximated as a unicycle, and the feet are represented by the unicycle wheels~\cite{8594277}. By sampling the continuous unicycle trajectory, it is possible to associate each unicycle pose to a time instant $t_{imp}$. This time instant can be considered as the time in which the swing foot impacts the ground. Once the impact time is defined, we decide to use it as a conditional variable to plan robot feasible footsteps, i.e. too fast/slow step duration and too long/short step length are avoided.
Once the footsteps are planned, the feet trajectory is obtained by cubic spline interpolation.
\par
The nominal footsteps position is also used to plan the nominal DCM trajectory. In details, the DCM is chosen so as to satisfy the following time evolution:
\begin{equation}
\label{eq:dcm_solution_ios}
\xi_i(t) = r^{zmp}_i + e ^{\frac{t}{b}} (\xi_i^{ios} - r^{zmp}_i),
\end{equation}
where $i$ indicates the $i$-th steps, $\xi_i^{ios}$ is the initial position of the DCM, $r^{zmp}_i$ is placed on the center of the stance foot and $t$ has to belong to the step domain $t \in [0, \; t^{step}_i]$.
Assuming that the final position of the DCM $\xi_{N-1}^{eos}$ coincides with the ZMP at last step, \eqref{eq:dcm_solution_ios} can be used to to define the following recursive algorithm for evaluating the DCM trajectory \cite{Englsberger2014}:
\begin{equation}
\label{eq:dcm_evaluation_ss}
\begin{cases}
\xi_{N-1}^{eos} = r_{N}^{zmp} \\
\xi_{i-1}^{eos} = \xi_{i}^{ios} = r_{i}^{zmp} + e^{-\frac{t^{step}_i}{b}} (\xi_i^{eos} - r_i^{zmp}) \\
\xi_i(t) = r^{zmp}_i + e ^{\frac{t}{b}} (\xi_i^{ios} - r^{zmp}_i).
\end{cases}
\end{equation}
As already pointed out in \cite{Englsberger2014}, the presented DCM planner has the main limitation of taking into account single support phases only. Indeed, by considering instantaneous transitions between two consecutive single support phases, the ZMP reference is discontinuous. This leads to the discontinuity of the external contact wrenches and consequentially of the desired joint torques.
The development of a DCM trajectory generator that handles non-instantaneous transitions between two single support phases becomes pivotal~\cite{Englsberger2014,Englsberger2019}. 
In the following, we decide to implement the solution proposed by Englsberger in \cite{Englsberger2014}. In order to guarantee a continuous ZMP trajectory, the desired DCM trajectory must belong at least to $C^1$ class. This can be easily guaranteed by smoothing two consecutive single support DCM trajectory by means of a third order polynomial function whose coefficients are chosen in order to satisfy the boundaries conditions, i.e. initial and final DCM position and velocities.

\subsection{Simplified model control}
The main goal of the simplified model control layer is to implement a control law based on the simplified model of the humanoid robot (see Section \ref{sec:simplified_models}) to stabilize the unstable DCM dynamics \eqref{eq:dcm_dynamics}. 
\par
To guarantee the tracking of the desired DCM trajectory the following control law is chosen:
\begin{equation}
\label{eq:instanteneous_dcm}
r^{vrp}_{des} = \xi_{ref} - b \dot{\xi}_{ref} + K^{\xi} (\xi - \xi_{ref}).
\end{equation}
\par
By using the control input \eqref{eq:instanteneous_dcm} in \eqref{eq:dcm_dynamics}, one has:
\begin{equation}
\label{eq:instanteneous_cl}
\dot{\tilde{\xi}} = \dot{\xi} - \dot{\xi}_{ref} = \frac{1}{b}(I_3 - K^{\xi})  \tilde{\xi}.
\end{equation}
The choice of  $K^{\xi}> I_3$ guarantees an asymptotically stable close loop error dynamics. 

\subsection{Whole-body QP control}
The whole-body QP control layer aims at the tracking of kinematic and force quantities. 
The proposed whole-body controller computes the desired joint torques  using the robot dynamics~\eqref{eq:robot_dynamic}, where the robot acceleration $\dot{\nu}$ and the  contact wrenches $f_k$ are set to desired \emph{starred} quantities, i.e.:
\begin{equation}
    \label{eq:forward_dynamics}
    \tau^{*} = B ^ \top \left(M(q) \dot{\nu}^{*} + h(q, \nu)  - \sum_{k = 1}^{n_c} J^\top_{\mathcal{C}_k}(q) f^{*}_k\right).
\end{equation}
\par
What follows evaluates desired generalized robot acceleration $\dot{\nu}^{*}$ and  contact wrenches $f^{*}_k$ that are compatible with the constraints acting on the system. Then, the  torques \eqref{eq:forward_dynamics} do not break the contacts the robot makes with the environment. 

\subsubsection{Desired generalized robot acceleration $\dot{\nu}^{*}$}
the desired generalized robot acceleration $\dot{\nu}^{*}$ are chosen to track the CoM acceleration, the torso orientation and the left and right feet position and orientation. To do so, we use a stack of task approach, which plays the role of a prioritized inverse kinematics in the acceleration space. The tracking of the feet trajectory and the CoM acceleration are considered as high priority tasks, while the torso orientation as well as a postural task are considered as low priority tasks. The control objective is achieved by designing the controller as a constrained optimization problem with
a cost function as:
\begin{equation}
\label{eq:torque_control_cost}
   H(\dot{\nu})= \frac{1}{2}  \left(  \norm{\dot{\omega}_{\mathcal{T}} ^ {des} - \dot{\omega}_{\mathcal{T}}}_{\Lambda_{\mathcal{T}}} ^2+ ~\norm{\ddot{s}\,^{des} - \ddot{s}}_{\Lambda _ s}^2 \right),
\end{equation}
where $\norm{a}_\Lambda$ indicates the weighted norm of the vector $a$. The tracking of the desired torso orientation is attempted via  the first term on the right hand side of \eqref{eq:torque_control_cost}, with the desired torso angular acceleration $\dot{\omega}^{des}_{\mathcal{T}}$. By choosing this desired acceleration properly, it is possible to guarantee almost-global stability and convergence of ${}^\mathcal{I}R _{\mathcal{T}}$ to ${}^\mathcal{I}R _{\mathcal{T}}^{ref}$ \cite{Olfati-Saber2000}, i.e.

\begin{equation}
\label{eq:rotational_pid_acceleration}
\begin{split}
    \dot{\omega}^{des}_\mathcal{T} &= \dot{\omega}^{ref} - c_0 \left(\hat{\omega} \prescript{\mathcal{I}}{}{R}_{\mathcal{T}} \prescript{\mathcal{I}}{}{R}_{\mathcal{T}} ^{ref^{\top}} - \prescript{\mathcal{I}}{}{R}_{\mathcal{T}} \prescript{\mathcal{I}}{}{R}_{\mathcal{T}} ^{ref^{\top}} \hat{\omega}^{ref}\right)^{\vee} \\
&- c_1 \left(\omega - \omega^{ref}\right) - c_2 \left(\prescript{\mathcal{I}}{}{R}_{\mathcal{T}} \prescript{\mathcal{I}}{}{R}_{\mathcal{T}} ^{ref^{\top}}\right) ^\vee,
\end{split}
\end{equation}
where $c_0$, $c_1$ and $c_2$ are positive numbers.
The postural task is encoded in \eqref{eq:torque_control_cost} thanks to the second term, which tends to penalise high  joint error by imposing
\begin{equation}
    \ddot{s} \,^{des} =  - K ^ d _s \dot{s} - K ^ p _s ( s  - s^ {ref}),
\end{equation}
where $K ^ p _s$ and $K ^ d _s$ are positive definite matrices.
Concerning the tracking of the feet and the CoM, we have:
\begin{equation}
\label{eq:cartesian_acceleration}
J_{\circ} \dot{\nu} = \dot{\mathrm{v}}^{des}_{\circ} - \dot{J}_{\circ} \nu \quad \quad  \circ = \{\mathcal{F}, \mathcal{C}\}.
\end{equation}
When the foot is in contact, the desired acceleration $\dot{\mathrm{v}}^{des}_{\mathcal{F}}$ is zero. During the swing phase, the angular part of $\dot{\mathrm{v}}^{des}_{\mathcal{F}}$ is given by \eqref{eq:rotational_pid_acceleration} where the subscript $\mathcal{T}$ is substitute with $\mathcal{F}$, while the linear part $\dot{v}^{des}$ is equal to:
\begin{equation}
    \label{eq:torque_feet_linear_pid}
    \dot{v}^{des}_{\mathcal{F}} =  \dot{v}\,^{ref}_{\mathcal{F}} - K^d _{x _{f}} (v_{\mathcal{F}} - v^{ref}_{\mathcal{F}}) - K^p _{x _{f}} (p_{\mathcal{F}} - p^{ref}_{\mathcal{F}}).
\end{equation}
Here the gains are again positive definite matrices. 
The CoM task is achieved by asking for a desired acceleration 
\begin{equation}
    \dot{\mathrm{v}}^{des}_{\mathcal{C}} = \frac{1}{b^2}\left(x - r_{des}^{vrp}\right)
\end{equation}
where the desired VRP $r_{des}^{vrp}$ is computed by the simplified model control layer -- see~\eqref{eq:instanteneous_dcm}.
It is also worth noting that when the CoM is considered in \eqref{eq:cartesian_acceleration}, the term $\dot{J}_{\mathcal{C}}$ is the linear part of the centroidal momentum matrix scaled by the robot mass \eqref{eq:centroidal_momentum_matrix}.
The above hierarchical control objectives can be summarized into a whole-body optimization problem:
\begin{IEEEeqnarray}{LCL}
\dot{\nu} ^* =& \argmin\limits_{\dot{\nu}}  \frac{1}{2} &  \left(  \norm{\dot{\omega}_{\mathcal{T}} ^ {des} - \dot{\omega}_{\mathcal{T}}}_{\Lambda_{\mathcal{T}}}^2 + ~\norm{\ddot{s}\,^{des} - \ddot{s}}_{\Lambda _ s} ^2 \right)  \label{eq:acceleration_QP}\\
&\text{s.t.} & \ddot{s} ^{des} =  - K ^ d _s \dot{s} - K ^ p _s ( s  - s^ {ref}) \nonumber \\
& & J_{\circ} \dot{\nu} = \dot{\mathrm{v}}^{des}_{\circ} - \dot{J}_{\circ} \nu \quad   \circ = \{\mathcal{LF}, \mathcal{RF}, \mathcal{C}\} .\nonumber 
\end{IEEEeqnarray}
The decision variable is the robot acceleration $\dot{\nu}$. Since  the  acceleration $\dot{\mathrm{v}}_\circ$ depends linearly on $\dot{\nu}$, the optimization problem can be cast into a QP one and solved efficiently.
\par

\subsubsection{Desired contact wrenches $f^{*}_k$} the desired contact wrenches are computed to track the desired centroidal momentum. To do so, we developed an optimization problem where the unknown variable $F$ is a stacked vector of contact forces $f_k$.
More precisely, the main objective of the optimization problem is to minimize the following cost function:
\begin{equation}
\label{eq:contact_wrench_mapping_cost}
    H(F) = \frac{1}{2} \left(\norm{\dot{h}^{des} - \dot{h}} _ { \Lambda _ H} ^2  + \norm{F} _ { \Lambda _ F} ^2 \right).
\end{equation}
The tracking of the desired centroidal momentum rate of change is attempted thanks to the first term of \eqref{eq:contact_wrench_mapping_cost}. The desired linear momentum rate of change is computed from the desired VRP -- see \eqref{eq:instanteneous_dcm} as
\begin{equation}
    \dot{h}_l^{des} = \frac{m}{b^2}\left(x - r_{des}^{vrp}\right).
\end{equation}
On the other hand, the desired rate of change of the angular momentum is computed by using the Eq.~\eqref{eq:centroidal_momentum_matrix_rate_of_change}, where $\dot{\nu}$ is substitute with the desired robot acceleration $\dot{\nu}^*$ evaluated by solving the optimization problem~\eqref{eq:acceleration_QP}.
The last term of ~\eqref{eq:contact_wrench_mapping_cost} is used to minimize the required contact wrenches.
The aforementioned optimization problem can be summarized as:
\begin{IEEEeqnarray}{LCL} 
\IEEEyesnumber \phantomsection
    \label{eq:fstar}
F^* = &\argmin_F & \frac{1}{2} \left(\norm{\dot{h}^{des} - \dot{h}} _ { \Lambda _ H} ^2  + \norm{F} _ { \Lambda _ F} ^2\right) \IEEEyessubnumber \\ 
& \mbox{s.t.} & C_f F \le 0 \IEEEyessubnumber \\ 
& &  \dot{h} = \sum_{k = 1}^{n_c}\begin{bmatrix}
      I_3 & 0_3 \\
      S(p_{\mathcal{C}_k} - p_{\mathcal{C}}) & I_3 
    \end{bmatrix} f_k + mg. \nonumber 
\end{IEEEeqnarray}
The decision variable is the contact wrenches $F$. Since the rate of change of the centroidal momentum depends linearly on $F$, the problem \eqref{eq:fstar} can be cast into a  QP one.
\par
Once the desired robot acceleration and the desired contact wrenches are computed, the desired joint torques can be easily evaluated by using the system dynamics \eqref{eq:forward_dynamics}.

\section{Step Adapter} 
\label{sec:stepAdaptation}

\begin{figure*}[!t]
  \centering
\includegraphics[scale=1]{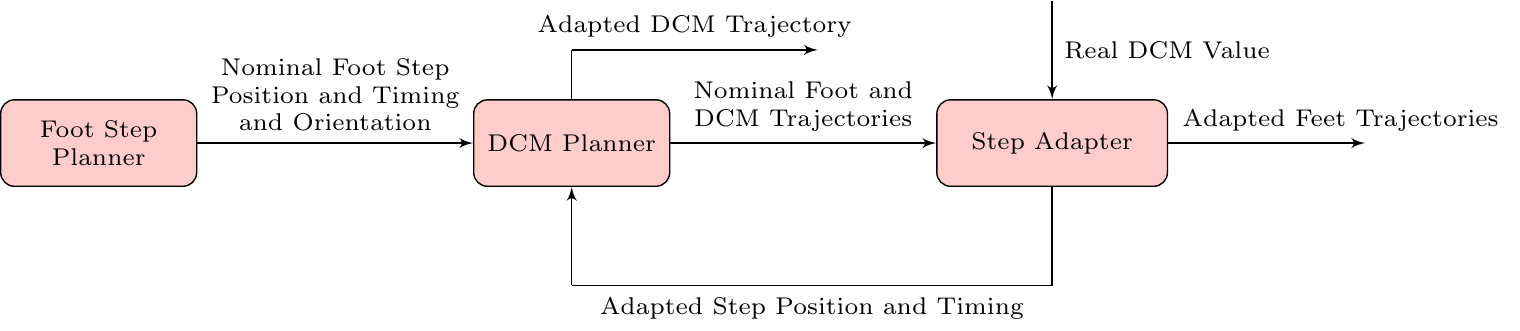}
  \caption{The trajectory optimization block in Fig.~\ref{fig:controller_architecture} is composed of three sub-modules: the \emph{footstep planner},  the \emph{DCM planner}, and the \emph{Step Adapter}.
  \label{fig:traj_opt}}
\end{figure*}

This section complements the architecture presented in Sec. \ref{sec:ARCHITETCURE} -- and shown in Fig. \ref{fig:controller_architecture} -- with a push recovery feature. As sketched in Fig. \ref{fig:traj_opt}, the \emph{trajectory optimisation} layer is augmented with a block called \emph{step adapter} to provide the architecture with the push recovery feature. 

More precisely, assuming the robot on a single support state, we propose below a step adjustment strategy that optimises the next step position and  timing while attempting to follow a nominal DCM trajectory. 
In particular, we define an exponential interpolation for the ZMP trajectory as: 
\begin{equation}
\label{eq:zmp_dynamics}
r^{zmp}(t)=A  e^{-\frac{t}{b} } + B,
\end{equation}
where $A ,B \in \mathbb{R}^2$ are chosen to satisfy the ZMP boundary conditions, i.e.  $r^{zmp}(0) =r^{zmp}_1$ $r^{zmp}(T) =r^{zmp}_2$, i.e.:
\begin{equation}
A =\frac{(r^{zmp}_2-r^{zmp}_1)\sigma}{1-\sigma}    \quad ,  \quad  B =\frac{r^{zmp}_1  - r^{zmp}_2 \sigma}{1- \sigma},
 \label{eq:ZMPcoefs} 
\end{equation}
with $T$ the duration of the single support phase, and

 \begin{equation}
 \label{eq:sigma}
\sigma:=e^{\frac{T}{b}}. 
\end{equation}
By substituting \eqref{eq:zmp_dynamics} into the DCM dynamics \eqref{eq:dcm_dynamics}, the following ordinary differential equation (ODE) holds:

\begin{equation}
\dot{\xi} - \frac{\xi}{b}=-\frac{1}{b} r^{zmp}(t)=- \frac{A}{b} e^{-\frac{t}{b}} - \frac{B}{b}.
\label{eq:DCMode}
\end{equation}
The solution to \eqref{eq:DCMode} writes:
\begin{equation}
\xi(t)=e^{ \int \frac{1}{b} dt} \left[  \int \left(- \frac{A}{b} e^{-\frac{t}{b} } -  \frac{B}{b}\right)  e^{\int -\frac{1}{b} dt}  dt+ C  \right],
 \label{eq:integralDCM}
\end{equation}
where $C\in \mathbb{R}^2$  is the vector of unknown coefficients that can be found by imposing the boundary conditions. Therefore, we can find these coefficients by solving the problem \eqref{eq:integralDCM} either as an initial value problem, namely
 \begin{equation}
 \label{eq:initial_value_problem}
\xi(0)=\xi_0=\frac{A}{2}+B+C_0, 
\end{equation}
or as a final value problem:
\begin{equation}
\label{eq:final_value_problem}
\xi(T)=\xi_T=\frac{A}{2}  e^{-\frac{T}{b}} +B+C_f  e^{\frac{T}{b}}.
\end{equation}

To find a DCM trajectory that satisfies both the initial and the final condition problems, the coefficient $C_0$ must equal $C_f$. Thus, by combining \eqref{eq:initial_value_problem} and \eqref{eq:final_value_problem}, one has:
\begin{equation}
\xi_0-\frac{A}{2}-B=\left(\xi_T-\frac{A}{2}  e^{-\frac{T}{b} } -B\right)  e^{-\frac{T}{b}}.
 \label{eq:constCoeef} 
\end{equation}
Now, define  $\delta=r^{zmp}_2-r^{zmp}_1$. Then, in view of \eqref{eq:sigma}, substituting (\ref{eq:ZMPcoefs}) to (\ref{eq:constCoeef}) yields: 
\begin{equation}
\xi_T-\frac{\delta}{2}(1+{\sigma})-r^{zmp}_1  + r^{zmp}_2  \sigma- \xi_0  \sigma=0.
 \label{eq:DCMConstraint0} 
\end{equation}
Let $r^{zmp}_T$ denote the ZMP position at the beginning of next step, and $\gamma_T=\xi_T-r^{zmp}_T$ the DCM offset. Therefore, straightforward calculations lead to:
\begin{equation}
\gamma_T+r^{zmp}_T+ \left(r^{zmp}_2- \xi_0 -\frac{\delta}{2} \right)\sigma=r^{zmp}_1 +\frac{\delta}{2}.
 \label{eq:DCMConstraint1} 
\end{equation}

The step adjustment problem can be formalized as a constrained optimization problem where the search variables are $\gamma_T$, $r^{zmp}_T$ and $\sigma$, and the cost function be properly defined to follow  nominal trajectory values. It is worth noting that the desired final DCM position and  step timing depend on $\gamma_T$ and $\sigma$, respectively. Meanwhile, $r^{zmp}_T$ is assumed to be at the center of the foot at the beginning of next step. Thus, we can assume this position to be considered as a target for the next footstep position. 

The following cost function is chosen to yield the desired gait values as close as possible to the nominal ones:
\begin{equation}
\begin{split}
J &=  \alpha_1  \norm{r^{zmp}_T - r^{zmp}_{T,nom}}^2 + \alpha_2  \norm{\gamma_{T} - \gamma_{nom}  }^2 \\
&+ \alpha_3  |\sigma -  e^{\frac{T_{nom}}{b}  } |^2,
 \label{eq:costStep} 
 \end{split}
\end{equation}
where $\alpha_1$, $\alpha_2$, $\alpha_3$ are positive numbers and the next ZMP position $r^{zmp}_{T,nom}$, step duration $T_{nom}$ and next DCM offset $\gamma_{nom}$ are obtained from the nominal trajectory generator.
\par
We also introduce the following inequality constraints: 
\begin{equation}
\begin{bmatrix}
 I_{2} &  0_{2 \times 1}   &0_{2}  \\
  -I_{2} &  0_{2 \times 1}   &0_{2}  \\
  0_{1 \times 2} & I_1  & 0_1  \\
  0_{1 \times 2} & -I_1  & 0_1  
\end{bmatrix}
\begin{bmatrix}
r^{zmp}_{T}  \\
 \sigma  \\
 \gamma_{T}  
\end{bmatrix}
\le
\begin{bmatrix}
  r^{zmp}_{T,max} \\
  -r^{zmp}_{T,min} \\
  \sigma_{max} \\ 
     -\sigma_{min}\\
\end{bmatrix},
\end{equation}
where $r^{zmp}_{T,max}, r^{zmp}_{T,min} \in \mathbb{R}^2$ and  $ \sigma_{max}, \sigma_{min} \in \mathbb{R}$. The inequality constraints are defined based on the maximum step length and minimum step duration that are related to the leg kinematic constraints and maximum realizable velocity, respectively. 
Finally, the relation described in \eqref{eq:DCMConstraint1} is treated as an equality constraint.
It is worth to notice that the linearity of \eqref{eq:DCMConstraint1} is ensured by the specific choice of \eqref{eq:zmp_dynamics}, and this choice also guarantees  the boundedness of the ZMP trajectory.
Since the cost function and the constraints depend quadratically and linearly on the unknown variables, all the above can be casted as a QP problem.
The QP problem is solved at each control cycle by substituting $\xi_0$ with the current DCM position and by progressively shrinking the single support duration as the robot completes the step.

Once the footstep position and the step timing are evaluated as solution to the QP problem, a new DCM trajectory is generated with the algorithm presented in section \ref{sec:trajectoryOptimization}.

Fig. \ref{fig:traj_opt} sketches the overall architecture for step recovery.
The foot step planner 
generates the nominal foot step position and timing, which are inputs of the DCM planner. 
The DCM planner generates the nominal DCM trajectory and foot position. The step adapter module combines 
nominal values (foot and DCM trajectories) and
measurements (actual DCM) to evaluate the adapted feet trajectories,   footstep position and timing. The former are as input of the whole-body QP control layer, while the latter are retrieved by the DCM planner to evaluate the adapted DCM trajectory.

\section{Results}
\label{sec:RESULTS}
In this section, we test the step adjustment approach alongside the architecture presented in Sec~\ref{sec:ARCHITETCURE}. Although the methodology presented above encompasses time-varying ZMP trajectories, the tests presented in this section are for fixed ZMPs during the single support phase.

More precisely, we present simulation results obtained with the implementation of the control architecture shown in Fig.~\ref{fig:controller_architecture}~\ref{fig:traj_opt}. 
We simulate the iCub humanoid robot \cite{Metta2010} using Gazebo simulator \cite{gazeboSim}.  Let us recall that iCub is $\SI{104}{\centi \meter}$ tall humanoid robot. Its weight is $\SI{33}{\kilo \gram}$ and  it has a foot length and width of  $\SI{19}{\centi \meter}$ and $\SI{9}{\centi \meter}$, respectively.

The architecture takes (in average) less than $\SI{3}{\milli \second}$ for evaluating its outputs: qpOASES \cite{Ferreau2014} was used to solve the QP problems. The  control modules run on the iCub head's computer, i.e. a 4-th generation Intel~\textsuperscript{\tiny\textregistered} Core i7 @ $\SI{1.7}{\giga \hertz}$. %

To validate the performances of the proposed architecture, we present two main push recovery experiments. 
First, the robot walks straight and an external disturbance acts on the pelvis to the lateral direction -- red arrows in Fig \ref{fig:footprintLeftPush}. Secondly, the robot follows a circular path and an external push is exerted on the pelvis as shown in Fig.~\ref{fig:dcmLeftPush2}. In both scenarios, the maximum straight velocity is $\SI{0.28}{\meter \per \second}$, and the external force has a magnitude of $\SI{150}{\newton}$  and lasts $\SI{0.05}{\second}$. The performances of the control architecture are shown in the accompanying video\protect\footnotemark.
\footnotetext{Video link: \url{https://youtu.be/DyNG8S6zznI}}
\subsection{Straight walking scenario}
\begin{figure*}[t]
  \centering
    \begin{subfigure}[b]{0.55\textwidth}
        \centering
        \includegraphics[height=4.8cm]{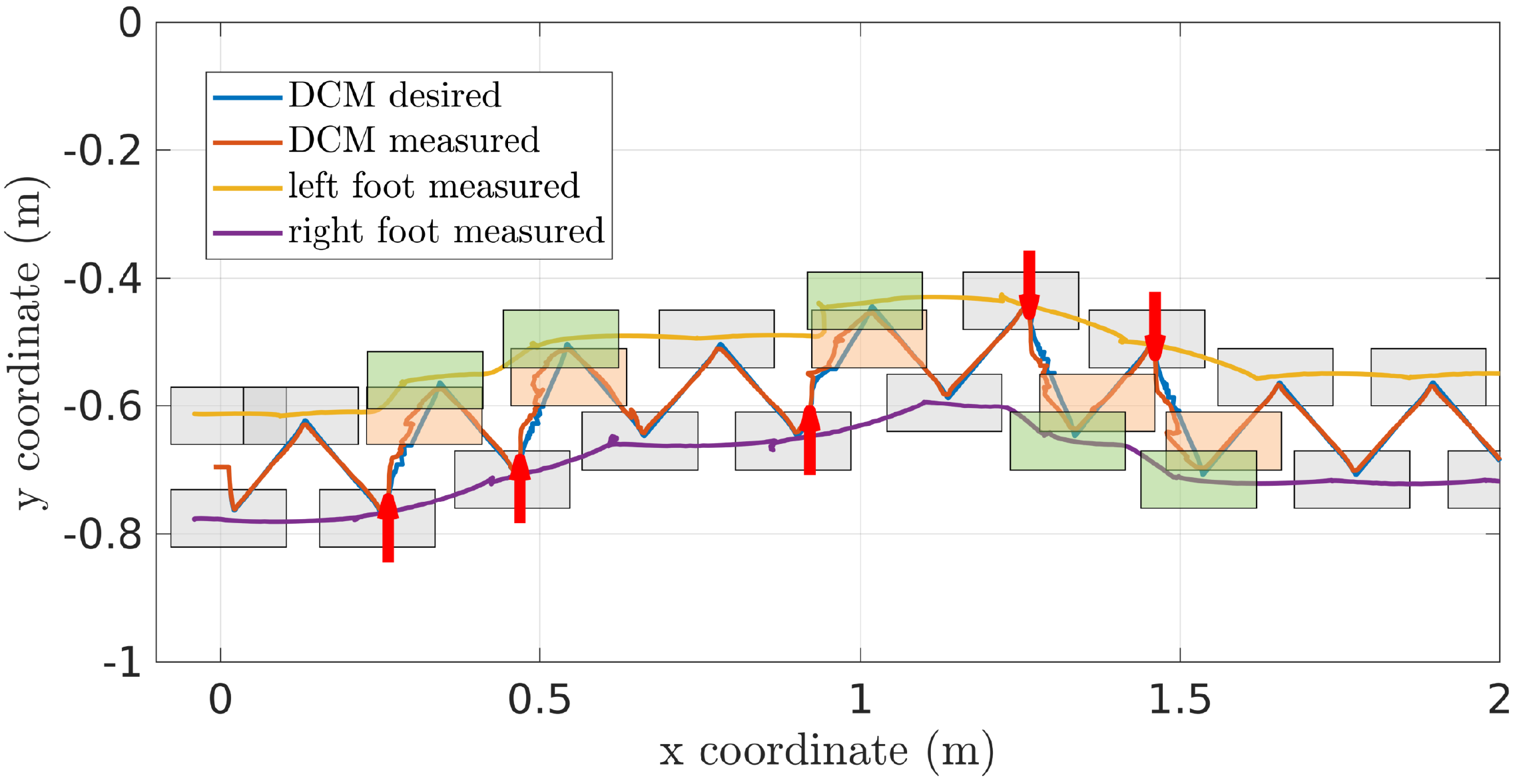}
       \caption{ \label{fig:footprintLeftPush}}
  \end{subfigure}
 \hfill
    \begin{subfigure}[b]{0.40\textwidth}
        \centering
    \includegraphics[height=4.8cm]{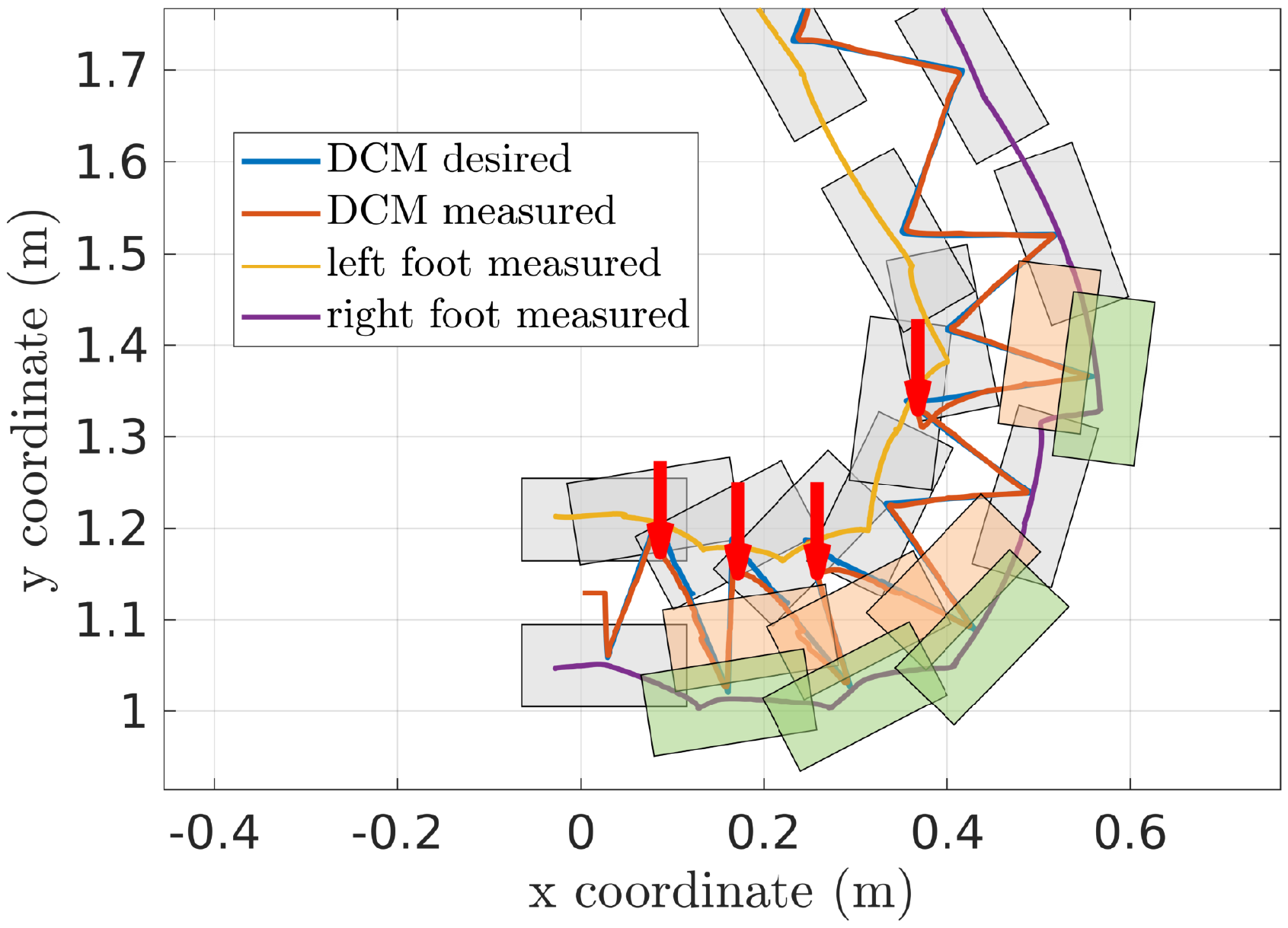}
   \caption{ \label{fig:dcmLeftPush2}}  
     \end{subfigure}
     \caption{Foot prints for straight (a) and circle path (b) push recovery scenario.  The push is indicated by red arrows.}
\end{figure*}

\begin{figure}[t]
\includegraphics[width=\columnwidth]{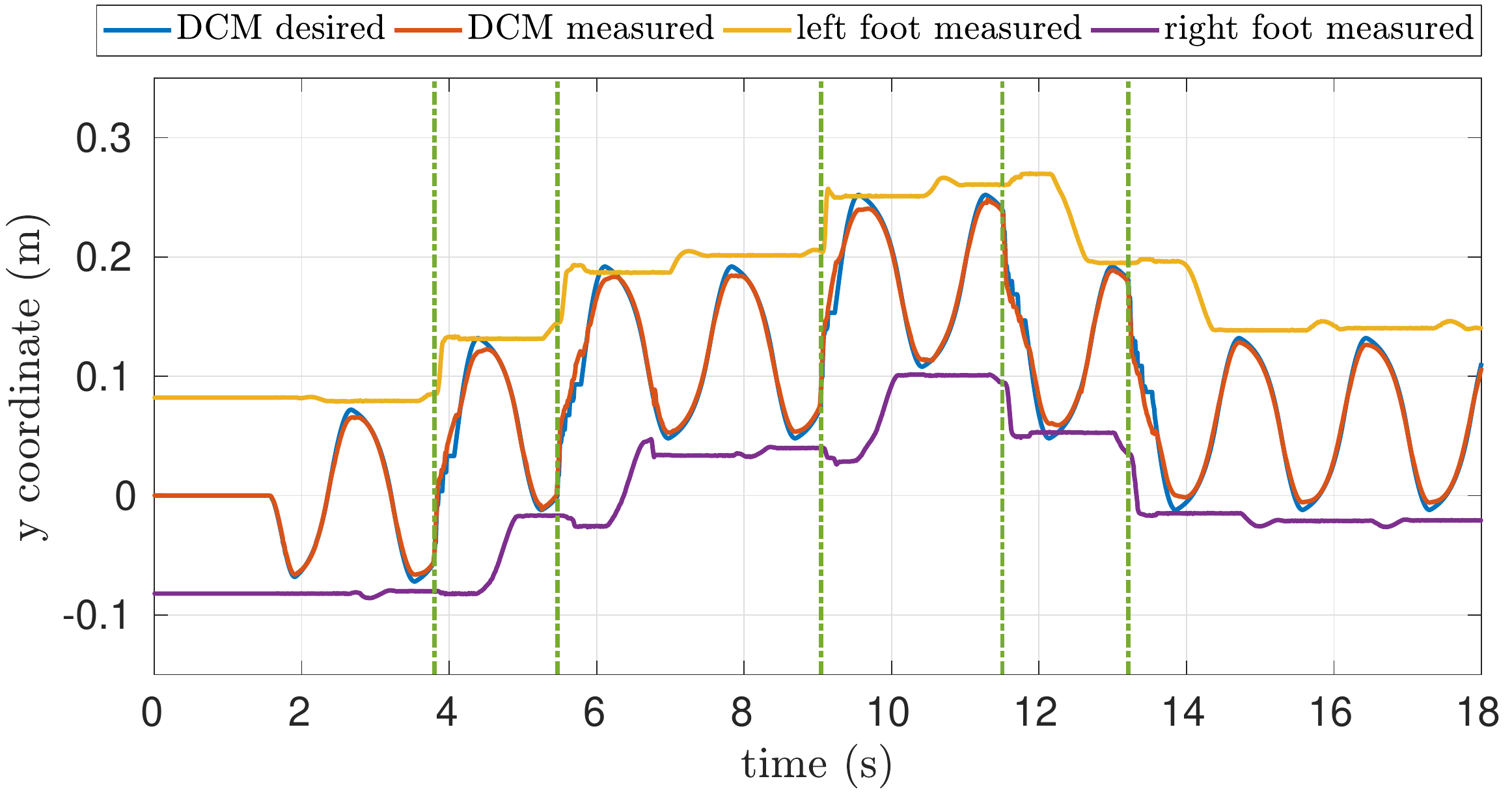}
 \caption{Trajectory of feet and DCM for lateral push recovery scenario. The instance that push  exerts  on the robot is indicated by green vertical lines.} 
  \label{fig:dcmLeftPush}
\end{figure}

\begin{figure}[t]
\includegraphics[width=\columnwidth]{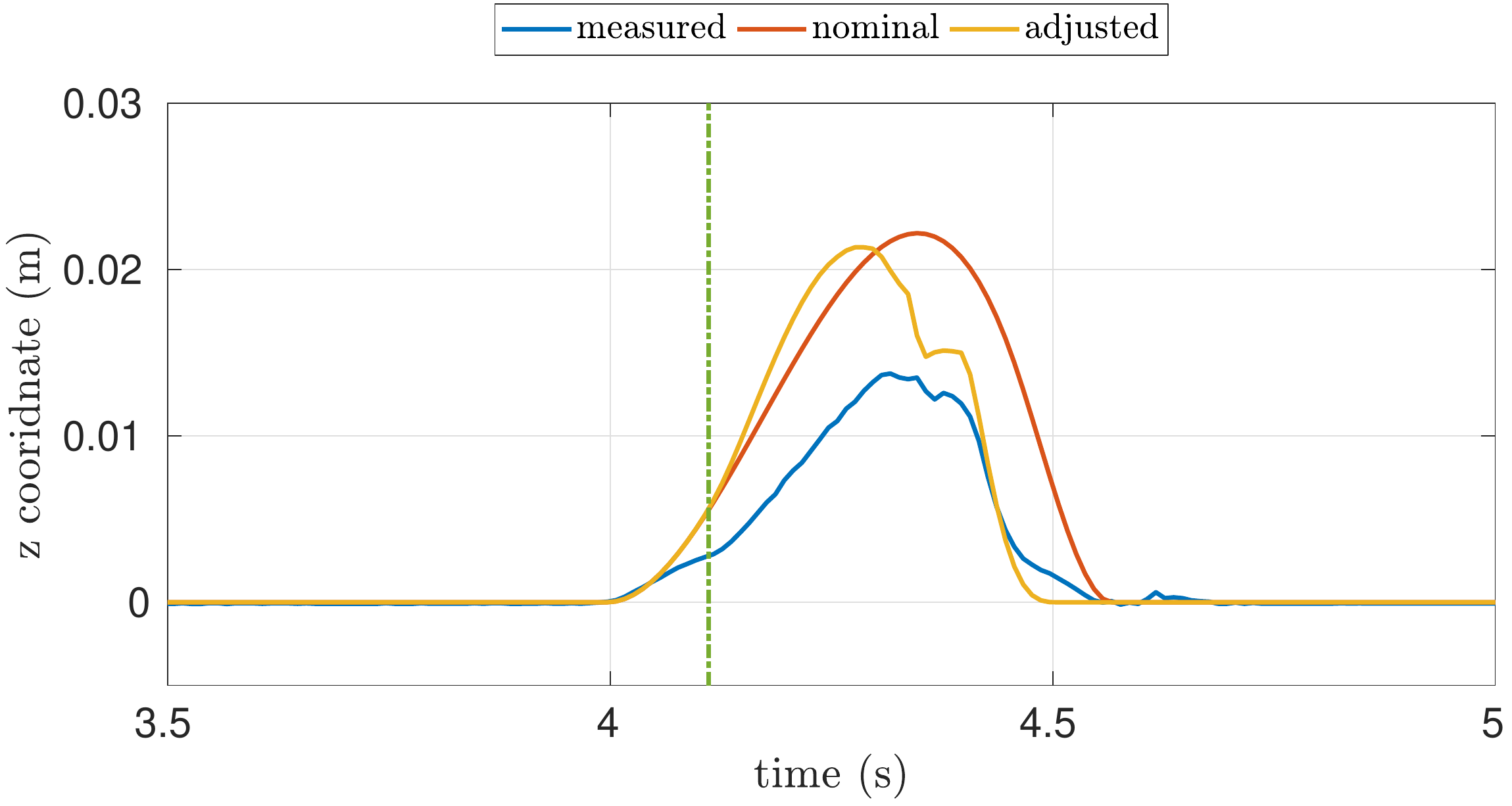}
 \caption{The z trajectory of the swing foot during step adjustment} 
  \label{fig:ZleftPush}
  \vskip -0.5cm
\end{figure}

Fig.~\ref{fig:dcmLeftPush} depicts the feet and DCM trajectory. The instant at which the disturbance acts on the robot is indicated by green vertical lines. The step adapter  compensates the disturbance effect by step timing and position optimization. The controller extends the step width with an average of $\SI{6}{\centi \meter}$, as shown in the Fig. \ref{fig:footprintLeftPush}. It reduces the step timing with an average of  $\SI{0.12}{\second}$ with respect to the nominal value of $\SI{0.53}{\second}$ and, as depicted in the Fig.~\ref{fig:ZleftPush}, the foot trajectory is adapted and step timing is decreased and the foot touches the ground sooner with respect to the nominal values.

Fig. \ref{fig:footprintLeftPush} shows in green and orange the position of the adapted footstep and the nominal one, respectively. If no external disturbances act on the robot, the position of the adjusted footsteps coincides with the nominal one, hand the footstep is represented by a gray rectangle. 

\subsection{Circular path following scenario}

In this scenario, the step adaptation algorithm compensates the external disturbance by extending the step length with an average of $\SI{6}{\centi \meter}$, and is applied for circular path walking.
 
 In Fig. \ref{fig:dcmLeftPush2}, the four disturbances exerted on the robot are indicated with red arrows. The nominal footprints corresponding to those instances are shown in orange. All  the adjusted footprints are shown in green. When the there is no disturbance acting on the robot, the footprints generated by the step adjustment algorithm coincides with the nominal one, thus are represented by using a gray rectangle. Since the push has components in both $x$ and $y$ coordinates, it activates a step adjustment both in sagittal and lateral directions.

\section{Conclusion and Future Work}
\label{sec:CONCLUSION}

This paper contributes towards robust bipedal locomotion of humanoid robots against external disturbances. 
The proposed approach fits control architecture where the DCM is planned beforehand, and allows a time-varying ZMP trajectory during the single support phase. Thus, applications of the proposed approach  enable the heel to toe motions. The step adapter presented in this paper optimises step position, timing,  and DCM  trajectory, all cast in  a  single  QP  problem. 
The proposed algorithm is employed in a three-layer DCM based controller architecture and validated on the simulated version of the iCub humanoid robot. During the tests, the robot undergoes perturbations in the form of external forces with magnitude of $\SI{150}{N}$, applied continuously for $\SI{0.05}{\second}$. Because of such disturbances, the presented algorithm determines suitable modifications to the pre-planned trajectories allowing to maintain balance. As a future work, we plan to validate this architecture also on the real robot.
\par
The efficiency of the presented algorithm relies on simplified DCM models. The cost of such choice is that the balancing capabilities of the robot may not be fully exploited, abd larger perturbations may require excessive modifications to the nominal trajectories. Hence, future work may consider also the robot angular momentum and its full kinematics.
Another possible improvement would consist in adopting the push recovery strategy also during the double support phase.
\addtolength{\textheight}{-11.7cm}   %
\bibliography{bibliography}

\begin{thebibliography}{10}
\providecommand{\url}[1]{#1}
\csname url@rmstyle\endcsname
\providecommand{\newblock}{\relax}
\providecommand{\bibinfo}[2]{#2}
\providecommand\BIBentrySTDinterwordspacing{\spaceskip=0pt\relax}
\providecommand\BIBentryALTinterwordstretchfactor{4}
\providecommand\BIBentryALTinterwordspacing{\spaceskip=\fontdimen2\font plus
\BIBentryALTinterwordstretchfactor\fontdimen3\font minus
  \fontdimen4\font\relax}
\providecommand\BIBforeignlanguage[2]{{%
\expandafter\ifx\csname l@#1\endcsname\relax
\typeout{** WARNING: IEEEtran.bst: No hyphenation pattern has been}%
\typeout{** loaded for the language `#1'. Using the pattern for}%
\typeout{** the default language instead.}%
\else
\language=\csname l@#1\endcsname
\fi
#2}}

\bibitem{feng2015optimization}
S.~Feng, E.~Whitman, X.~Xinjilefu, and C.~G. Atkeson, ``{Optimization-based
  Full Body Control for the DARPA Robotics Challenge},'' \emph{J. F. Robot.},
  vol.~32, no.~2, pp. 293--312, 2015.

\bibitem{dai2014whole}
H.~Dai, A.~Valenzuela, and R.~Tedrake, ``{Whole-body motion planning with
  centroidal dynamics and full kinematics},'' in \emph{2014 IEEE-RAS Int. Conf.
  Humanoid Robot.}\hskip 1em plus 0.5em minus 0.4em\relax IEEE, 2014, pp.
  295--302.

\bibitem{herzog2015trajectory}
A.~Herzog, N.~Rotella, S.~Schaal, and L.~Righetti, ``{Trajectory generation for
  multi-contact momentum control},'' in \emph{Humanoid Robot. (Humanoids), 2015
  IEEE-RAS 15th Int. Conf.}\hskip 1em plus 0.5em minus 0.4em\relax IEEE, 2015,
  pp. 874--880.

\bibitem{PascalHandbook}
P.~Morin and C.~Samson, \emph{{Handbook of Robotics}}.\hskip 1em plus 0.5em
  minus 0.4em\relax Springer, 2008, ch. Motion con, pp. 799--826.

\bibitem{flavigne2010reactive}
D.~Flavigne, J.~Pettr{\'{e}}e, K.~Mombaur, J.-P. Laumond, and Others,
  ``{Reactive synthesizing of human locomotion combining nonholonomic and
  holonomic behaviors},'' in \emph{Biomed. Robot. Biomechatronics (BioRob),
  2010 3rd IEEE RAS EMBS Int. Conf.}\hskip 1em plus 0.5em minus 0.4em\relax
  IEEE, 2010, pp. 632--637.

\bibitem{8594277}
S.~Dafarra, G.~Nava, M.~Charbonneau, N.~Guedelha, F.~Andradel, S.~Traversaro,
  L.~Fiorio, F.~Romano, F.~Nori, G.~Metta, and D.~Pucci, ``{A Control
  Architecture with Online Predictive Planning for Position and Torque
  Controlled Walking of Humanoid Robots},'' in \emph{2018 IEEE/RSJ Int. Conf.
  Intell. Robot. Syst.}, 2018, pp. 1--9.

\bibitem{Kajita2001}
S.~Kajita, F.~Kanehiro, K.~Kaneko, K.~Yokoi, and H.~Hirukawa, ``{The 3D linear
  inverted pendulum model: a simple modeling for biped walking pattern
  generation},'' \emph{Proc. 2001 IEEE/RSJ Int. Conf. Intell. Robot. Syst.},
  no. October 2016, pp. 239--246, 2001.

\bibitem{Pratt2006}
J.~Pratt, J.~Carff, S.~Drakunov, and A.~Goswami, ``{Capture point: A step
  toward humanoid push recovery},'' in \emph{Proc. 2006 6th IEEE-RAS Int. Conf.
  Humanoid Robot. Humanoids}, 2006, pp. 200--207.

\bibitem{Hof2008}
A.~L. Hof, ``{The 'extrapolated center of mass' concept suggests a simple
  control of balance in walking},'' \emph{Hum. Mov. Sci.}, 2008.

\bibitem{Vukobratovic1969}
M.~Vukobratovic and D.~Juricic, ``{Contribution to the Synthesis of Biped
  Gait},'' \emph{IEEE Trans. Biomed. Eng.}, vol. BME-16, no.~1, pp. 1--6, 1969.

\bibitem{Englsberger2015}
J.~Englsberger, C.~Ott, and A.~Albu-Sch{\"{a}}ffer, ``{Three-Dimensional
  Bipedal Walking Control Based on Divergent Component of Motion},'' \emph{IEEE
  Trans. Robot.}, vol.~31, no.~2, pp. 355--368, 2015.

\bibitem{Stephens2010}
B.~J. Stephens and C.~G. Atkeson, ``{Dynamic Balance Force Control for
  compliant humanoid robots},'' in \emph{Intell. Robot. Syst. (IROS), 2010
  IEEE/RSJ Int. Conf.}, 2010, pp. 1248--1255.

\bibitem{nava16}
G.~Nava, F.~Romano, F.~Nori, and D.~Pucci, ``{Stability Analysis and Design of
  Momentum-based Controllers for Humanoid Robots},'' \emph{Intell. Robot. Syst.
  2016. IEEE Int. Conf.}, 2016.

\bibitem{feng2016robust}
S.~Feng, X.~Xinjilefu, C.~G. Atkeson, and J.~Kim, ``Robust dynamic walking
  using online foot step optimization,'' in \emph{2016 IEEE/RSJ International
  Conference on Intelligent Robots and Systems (IROS)}.\hskip 1em plus 0.5em
  minus 0.4em\relax IEEE, 2016, pp. 5373--5378.

\bibitem{Shafiee-Ashtiani2017}
M.~Shafiee-Ashtiani, A.~Yousefi-Koma, and M.~Shariat-Panahi, ``{Robust bipedal
  locomotion control based on model predictive control and divergent component
  of motion},'' in \emph{Proc. - IEEE Int. Conf. Robot. Autom.}\hskip 1em plus
  0.5em minus 0.4em\relax IEEE, may 2017, pp. 3505--3510.

\bibitem{joe2018balance}
H.-M. Joe and J.-H. Oh, ``Balance recovery through model predictive control
  based on capture point dynamics for biped walking robot,'' \emph{Robotics and
  Autonomous Systems}, vol. 105, pp. 1--10, 2018.

\bibitem{khadiv2016step}
M.~Khadiv, A.~Herzog, S.~A.~A. Moosavian, and L.~Righetti, ``Step timing
  adjustment: A step toward generating robust gaits,'' in \emph{2016 IEEE-RAS
  16th International Conference on Humanoid Robots (Humanoids)}.\hskip 1em plus
  0.5em minus 0.4em\relax IEEE, 2016, pp. 35--42.

\bibitem{griffin2017walking}
R.~J. Griffin, G.~Wiedebach, S.~Bertrand, A.~Leonessa, and J.~Pratt, ``Walking
  stabilization using step timing and location adjustment on the humanoid
  robot, atlas,'' in \emph{2017 IEEE/RSJ International Conference on
  Intelligent Robots and Systems (IROS)}.\hskip 1em plus 0.5em minus
  0.4em\relax IEEE, 2017, pp. 667--673.

\bibitem{jeong2019robust1}
H.~Jeong, I.~Lee, J.~Oh, K.~K. Lee, and J.-H. Oh, ``A robust walking controller
  based on online optimization of ankle, hip, and stepping strategies,''
  \emph{IEEE Transactions on Robotics}, 2019.

\bibitem{kuo2002energetics}
A.~D. Kuo, ``Energetics of actively powered locomotion using the simplest
  walking model,'' \emph{Journal of biomechanical engineering}, vol. 124,
  no.~1, pp. 113--120, 2002.

\bibitem{ogura2006human}
Y.~Ogura, K.~Shimomura, H.~Kondo, A.~Morishima, T.~Okubo, S.~Momoki, H.-o. Lim,
  and A.~Takanishi, ``Human-like walking with knee stretched, heel-contact and
  toe-off motion by a humanoid robot,'' in \emph{2006 IEEE/RSJ International
  Conference on Intelligent Robots and Systems}.\hskip 1em plus 0.5em minus
  0.4em\relax IEEE, 2006, pp. 3976--3981.

\bibitem{Englsberger2014}
J.~Englsberger, T.~Koolen, S.~Bertrand, J.~Pratt, C.~Ott, and
  A.~Albu-Sch{\"{a}}ffer, ``{Trajectory generation for continuous leg forces
  during double support and heel-to-toe shift based on divergent component of
  motion},'' in \emph{IEEE Int. Conf. Intell. Robot. Syst.}, 2014, pp.
  4022--4029.

\bibitem{romualdi2018benchmarking}
G.~Romualdi, S.~Dafarra, Y.~Hu, and D.~Pucci, ``A benchmarking of dcm based
  architectures for position and velocity controlled walking of humanoid
  robots,'' in \emph{2018 IEEE-RAS 18th International Conference on Humanoid
  Robots (Humanoids)}.\hskip 1em plus 0.5em minus 0.4em\relax IEEE, 2018, pp.
  1--9.

\bibitem{Natale2017}
L.~Natale, C.~Bartolozzi, D.~Pucci, A.~Wykowska, and G.~Metta, ``{iCub: The
  not-yet-finished story of building a robot child},'' \emph{Sci. Robot.},
  2017.

\bibitem{Marsden2010}
J.~E. Marsden and T.~S. Ratiu, \emph{{Introduction to Mechanics and Symmetry: A
  Basic Exposition of Classical Mechanical Systems}}.\hskip 1em plus 0.5em
  minus 0.4em\relax Springer Publishing Company, Incorporated, 2010.

\bibitem{orin2008centroidal}
D.~E. Orin and A.~Goswami, ``Centroidal momentum matrix of a humanoid robot:
  Structure and properties,'' in \emph{2008 IEEE/RSJ International Conference
  on Intelligent Robots and Systems}.\hskip 1em plus 0.5em minus 0.4em\relax
  IEEE, 2008, pp. 653--659.

\bibitem{Englsberger2019}
J.~Englsberger, G.~Mesesan, C.~Ott, and A.~Albu-Schaffer, ``{DCM-Based Gait
  Generation for Walking on Moving Support Surfaces},'' 2019.

\bibitem{Olfati-Saber2000}
R.~Olfati-Saber, ``{Nonlinear Control of Underactuated Mechanical Systems with
  Application to Robotics and Aerospace Vehicles},'' Ph.D. dissertation, 2000.

\bibitem{Metta2010}
G.~Metta, L.~Natale, F.~Nori, G.~Sandini, D.~Vernon, L.~Fadiga, C.~von Hofsten,
  K.~Rosander, M.~Lopes, J.~Santos-Victor, A.~Bernardino, and L.~Montesano,
  ``{The iCub humanoid robot: An open-systems platform for research in
  cognitive development},'' \emph{Neural Networks}, 2010.

\bibitem{gazeboSim}
N.~{Koenig} and A.~{Howard}, ``Design and use paradigms for gazebo, an
  open-source multi-robot simulator,'' in \emph{2004 IEEE/RSJ International
  Conference on Intelligent Robots and Systems (IROS) (IEEE Cat.
  No.04CH37566)}, vol.~3, Sep. 2004, pp. 2149--2154 vol.3.

\bibitem{Ferreau2014}
H.~J. Ferreau, C.~Kirches, A.~Potschka, H.~G. Bock, and M.~Diehl,
  ``{{\{}qpOASES{\}}: A parametric active-set algorithm for quadratic
  programming},'' \emph{Math. Program. Comput.}, vol.~6, no.~4, pp. 327--363,
  2014.

\end{thebibliography}
\bibliographystyle{IEEEtran}

\end{document}